\documentclass[final]{article}

\usepackage[nonatbib]{neurips_2024}

\usepackage[utf8]{inputenc} % allow utf-8 input
\usepackage[T1]{fontenc}    % use 8-bit T1 fonts
\usepackage{hyperref}       % hyperlinks
\usepackage{url}            % simple URL typesetting
\usepackage{booktabs}       % professional-quality tables
\usepackage{amsfonts}       % blackboard math symbols
\usepackage{nicefrac}       % compact symbols for 1/2, etc.
\usepackage{microtype}      % microtypography
\usepackage{xcolor}         % colors

\usepackage{listings}
\usepackage{amsmath}
\usepackage{graphicx}
\usepackage[
    backend=biber,
    natbib=true,
]{biblatex}
\title{Introduction to Sequence Modeling with Transformers}

\author{%
  Joni-Kristian K{\"a}m{\"a}r{\"a}inen\thanks{See \url{https://webpages.tuni.fi/vision/public_pages/JoniKamarainen/}}\\
  Department of Computing Sciences\\
  Tampere University
}

\begin{document}

\maketitle

\begin{abstract}
  Understanding the transformer architecture and its workings is
  essential for machine learning (ML) engineers. However, truly
  understanding the transformer architecture can be demanding, even if
  you have a solid background in machine learning or deep
  learning. The main working horse is attention, which yields to the
  transformer encoder-decoder structure. However, putting attention
  aside leaves several programming components that are easy to
  implement but whose role for the whole is unclear. These components
  are 'tokenization', 'embedding' ('un-embedding'), 'masking', 'positional encoding', and 'padding'. The focus of this work is on understanding
  them. To keep things simple, the understanding is built
  incrementally by adding components one by one, and after each step
  investigating what is doable and what is undoable with the current
  model. Simple sequences of zeros (0) and ones (1) are used to study
  the workings of each step.
\end{abstract}

\begin{center}
\texttt{Jupyter notebook:} \url{https://github.com/kamarain/transformer_intro}
\end{center}

\section{Background}
If you already know \textit{machine learning} and
\textit{sequence modeling} well, then this section can be skipped. The
purpose of this section is to 
cover the terminology and concepts used in the work.

\paragraph{Supervised machine learning.}
The supervised machine learning problem is to find a model $f$ that
maps inputs $x$ to outputs $y$
\begin{equation}
  y = f(x) \enspace .
  \label{eq:supervisedML}
\end{equation}
The conventional methods for the ML problem are regression and classification.
Before the recent deep learning methods,
the most popular neural ML model was Multilayer Perceptron (MLP). The
fundamental ideas and methods of conventional ML including MLPs are
well covered in Bishop's ML book from 2006~\cite{MLBook}
(available online).

Using the deep learning terminology, MLP is a neural network of
\textit{fully connected layers} and \textit{non-linear activation
  functions} (such as ReLU or Sigmoid) between them. Through the
layers, MLP maps inputs $x$ to outputs $y$. Each neuron (\textit{unit}
or \textit{kernel}) has weights which are the learnable
parameters $\theta$ of the network. Network training is based on
adjusting the weights so that some error (\textit{loss function})
is minimized. The typical loss function for regression is the Mean
Squared Error (MSE) and for classification Cross Entropy (CE). The MSE
loss is defined
\[
\mathcal{L}_{MSE} = \sum_{i=1}^{K} (y_i-f_\theta (x))^2 \enspace .
\]

The loss is computed for a set of $K$ input-output pairs, which is
the \textit{training data}. Formally, the optimization is defined
as
\[
\theta = \arg\min_{\theta} \mathcal{L}_{MSE} \enspace ,
\]
and the standard technique for optimization is the Stochastic Gradient
Descent (SGD) algorithm. The process of passing the loss gradient
through the network is called \textit{error backpropagation}.

A Torch-style code block of a two-layer MLP network:
\begin{lstlisting}
.init(self, input_dim, hidden_size, output_dim):
        self.dense1 = torch.nn.Linear(input_dim, hidden_size)
        self.dense2 = torch.nn.Linear(hidden_size, hidden_size)
        self.output = torch.nn.Linear(hidden_size, output_dim)

.forward(self, x):
    x = self.dense1(x)
    x = torch.nn.functional.sigmoid(x)
    x = self.dense2(x)
    x = torch.nn.functional.sigmoid(x)
    y_pred = self.output(x)
    return y_pred
\end{lstlisting}

\paragraph{Sequence modeling.}
A problem domain where sequence modeling is used is time series
forecasting. For example, forecasting future stock prices is a time
series forecasting application. The problem can be defined as the
problem of predicting the future values 
$\hat{\mathbf{Y}} = \left\{ \hat{x}_N, \hat{x}_{N+1}, \ldots, \hat{x}_{N+M} \right\}$of a process
from which past values $\mathbf{X} = \left\{ x_0, x_{1}, \ldots, x_{N-1} \right\}$ are known.
The hat symbol $\hat{\cdot}$ is used to distinguish between the true
(ground truth) and predicted values.

An ML-compliant definition is to find the model $f_\theta$ so that
\[
\mathbf{Y} = f_\theta (\mathbf{X}) 
\]
and MLP can be used to find a model. The problem with the MLP model is
that the length of the sequences $\mathbf{X}$ and $\mathbf{Y}$ can vary,
but MLP must have a fixed number of inputs and outputs. For iterative
forecasting, the number of outputs can be set to one, and the number
of inputs can be 'big enough', and if the input sequence is
shorter than the maximum limit, \textit{padding inputs} are added.
However, such a network grows rapidly for an increasing number of
inputs and soon becomes useless for practical applications where only
a limited number of training samples are available.

The number of inputs can be reduced, only one past example being the
limit. In this case, the MLP learns the mapping
\[
x_{n+1} = f_\theta (x_{n}) \enspace ,
\]
but this naïve approach does not work very well. For example, it
cannot model sinusoidal signals since each value except 0 and 1
appears in two ’contexts'; the raising and falling edges of the wave.
At the same time, it is well known that some problems need to address
both short and long-horizon dependencies to predict the next value.
This problem can be solved by Recurrent Neural Networks (RNNs), in
theory, and by Transformers, in practice.

\paragraph{Sequence-to-sequence (Seq2Seq) modeling.}
In the deep learning literature, Seq2Seq models are discussed in the context of
Recurrent Neural Networks (RNNs). The main difference between the MLP
and RNN models is that RNNs maintain a system 'state' $h_t$ that is
passed forward to the next prediction. The RNN model can be
defined as
\[
x_{n+1},h_{n+1} = f_\theta (x_{n},h_{n}) \enspace .
\]

Theoretically, RNNs can have infinitely long 'memory', but in practice,
that would mean that gradient information is maintained over a
long period of time, which is difficult with limited precision of computing 
and training time. 

Many RNN architectures have been proposed, and one of the most popular
is Long Short-Term Memory (LSTM)~\cite{LSTM}.

\paragraph{Transformer architecture.}
To understand the design choices of the Transformer architecture, it
can be beneficial to learn about the progression of RNNs in machine
translation where one language is mapped to another, such as
English-to-French. This Seq2Seq problem is more general than the
time series forecasting since the type of input and output are
different (English vs. French vocabulary).

An excellent work that describing the principle idea of using RNNs
(LSTM) in machine translation is~\cite{Sutskever-2014-neurips}.
An encoder LSTM encodes the input sequence of $N$ inputs into a single summary vector.
In practice, the summary vector is the last state $h_n$ of the LSTM. Then
the state vector is passed to the decoder LSTM that starts to generate
output. The starts and ends of both sequences are encoded using
special symbols: Start-of-Sequence (SOS) and End-of-Sequence (EOS).

A similar structure was used in~\cite{Cho-2014-emnlp}, which continues the story of~\cite{Sutskever-2014-neurips}. The authors noted that the LSTM encoder-decoder
structure quickly forgets history, and propose a mechanism that uses
information from all past samples to infer the next prediction(s).
This mechanism is \textit{attention}. In their model,  all past state
vectors are stored, multiplied by \textit{attention weights}, and
summed for the decoder. Similarly, the decoder can use its previous
outputs via \textit{self-attention}. This approach can use all input samples in
inference. Them main bottleneck is the sequential process of RNNs
which makes training and inference slow. This attention mechanism is
often referred to as \textit{Bahdanau attention}.

Finally, after the two previous and influential works, the seminal
Transformer paper ``Attention is All You Need'' is much easier to
comprehend with all its flavors~\cite{transformer}. The most important
idea is to avoid the sequential processing of RNNs. This is achieved
by multiplicative attention that exploits the neural computation
library (Torch, Keras) functions ability to automatically forward any
number of inputs and backward the gradients from the loss function. In
many ways, Transformer is more like clever software engineering than
new AI theory.

\section{Methods}

In the following sections, we incrementally add the essential
components to the plain transformer and demonstrate how its
capabilities gradually improve in Seq2Seq modeling. In addition, implementation details related to the \texttt{torch.nn.Transformer} module are dicussed since some of them are not well documented.

\subsection{Plain Transformer}
\label{sec:PlainTransformer}

At first, we investigate a neural model with Transformer as the only processing element.
This model is referred to as \texttt{PlainTransformer}. Our reference implementation is \texttt{torch.nn.Transformer} which claims to follow the original Transformer paper~\cite{transformer}. The reference implementation hides details of internal processing steps, and is thus straightforward to use.

\paragraph{Normalization before attention.} The main difference between our implementation and  the original work is the feature normalization. In our case, the normalization is done \textit{before} the multiheaded attention. Normalization before the attention stabilizes training and removes the need for heat-up epochs with
increasing learning rates. The before normalization was proposed
in~\cite{Xiong-2020-icml} and verified by our preliminary experiments. In \texttt{torch.nn.Transformer}, the normalization is changed by
setting the Transformer parameter \texttt{norm\_first = True} (default: False).

Python code for \texttt{PlainTransformer}:
\begin{lstlisting}
class PlainTransformer(nn.Module):
    def __init__(
        self,
        d_model,
        nhead,
        num_encoder_layers,
        num_decoder_layers,
        dim_feedforward,
        dropout_p,
        layer_norm_eps
    ):
        super().__init__()

        # Transformer initialized with user specs
        self.transformer = nn.Transformer(
            d_model = d_model,
            nhead = nhead,
            num_encoder_layers = num_encoder_layers,
            num_decoder_layers = num_decoder_layers,
            dim_feedforward = dim_feedforward,
            dropout = dropout_p,
            layer_norm_eps = layer_norm_eps,
            norm_first = True
        )

    def forward(
        self,
        src,
        tgt,
    ):
        # Transformer assumes that src & tgt structure is (seq_length, batch_num, feat_dim)
        out = self.transformer(src, tgt)

        return out
\end{lstlisting}

\paragraph{PlainTransformer.}
An instance of the \texttt{PlainTransformer} model is constructed
with a minimal number of learnable parameters:
\begin{lstlisting}
model = SimplestTransformer(d_model = 1, nhead = 1, num_encoder_layers = 1,
                            num_decoder_layers = 1, dim_feedforward = 8,
                            dropout_p = 0.1, layer_norm_eps = 1e-05)
\end{lstlisting}

For some learning capacity, the Transformer feedforward layers have
8 linear units.
With one unit the number of
learnable parameters is 46, and 88 with 8 units. The other learnable parameters are related to the normalization layers and
other learnable operations hidden inside the reference implementation.

\paragraph{Training sequences.} The limits of the plain transformer model can be exemplified with simple data where binary inputs are repeated in the outputs:
\begin{displaymath}
  \begin{split}
    \mathbf{X}: 0,0,0,0 \rightarrow \mathbf{Y}: 0,0,0,0\\
    \mathbf{X}: 1,1,1,1 \rightarrow \mathbf{Y}: 1,1,1,1
  \end{split}
\end{displaymath}
Despite that there are only two training samples, 50 of each were generated (100 samples in total).

\paragraph{Results.} PlainTransformer was trained 300 epochs using
the Adam optimizer with the learning rate 0.01. The final
training loss was 0.25.

The trained model produced the following outputs:
\begin{displaymath}
  \begin{split}
    \mathbf{X}: 0,0,0,0 \rightarrow \hat{\mathbf{Y}}: 0.5,0.5,0.5,0.5\\
    \mathbf{X}: 1,1,1,1 \rightarrow \hat{\mathbf{Y}}: 0.5,0.5,0.5,0.5\\
  \end{split}
\end{displaymath}

\paragraph{Findings.} The transformer encoder is designed to
``correlate'' each input sample with all other
samples. The encoder outputs still represent the original samples, but now each of them also carries information about the whole sequence. The last encoder step is linear layer mapping. The purpose of the linear mapping is to transform encoder feature space features
to decoder feature space features. The feature spaces and mappings are learned during training.

The decoder performs a similar correlation between but with both input and so far produced output samples. Therefore, input to the decoder linear mapping is aware of all samples.  The purpose of the decoder linear mapping is to transfer
decoder feature space vectors to prediction feature space. This mapping is followed by a suitable loss function.

In \texttt{PlainTransformer} training, the decoder outputs were
directly used as the predictions. There are no non-linear mappings
in the processing pipeline, and therefore the plain transformer converges to
produce the output sample mean that is the single value minimizing error.

\subsection{Token embedding and un-embedding}

The results with \texttt{PlainTransformer} show
that the Transformer inputs and outputs should be higher dimensional
\textit{embedding} vectors which the Transformer can alter to
produce informative representations of the input and
output sequences.

As the solution, the \texttt{torch.Embedding()} module constructs a
lookup table with learnable embedding vectors. Inputs are unique 'tokens' denoted by integer values from 0 to N, and the Embedding module converts them to D-dimensional vectors that carry more information.

The following changes are made to the \texttt{PlainTransformer} code
to change it to \texttt{TokenTransformer}:
\begin{lstlisting}
class TokenTransformer(nn.Module):
...

        .__init__(...)
        ...
        # Token embedding layer - this takes care of converting integer ids to vectors
        self.embedding = nn.Embedding(num_tokens, d_model)

        # Token "unembedding" to one-hot encoded token vector
        self.unembedding = nn.Linear(d_model, num_tokens)
        ...
        
       .forward(...)
       ...
        # Note: src & tgt default size is (seq_length, batch_num, feat_dim)

        # Token embedding
        src = self.embedding(src) * math.sqrt(self.d_model)
        tgt = self.embedding(tgt) * math.sqrt(self.d_model)        

        # Transformer output
        out = self.transformer(src, tgt)
        out = self.unembedding(out)
        
        return out
\end{lstlisting}

In the above code, the embedding vectors are multiplied by the square
root of the feature vector dimension ($\sqrt{D}$). That should improve convergence in training~\cite{transformer}. With feature embeddings, the number of learnable
parameters of the embedding increases from 88 to 1332.

\paragraph{Tokenization.}
'Tokens' are used to represent input and output sequences. The natural choices
are
\begin{itemize}
\item 0 for '0' and
\item 1 for '1' .
\end{itemize}

In addition, to properly initialize and finish output generation,
two special tokens, Start-of-Sequence (SOS) and
End-of-Sequence (EOS), are added to the vocabulary:
\begin{itemize}
\item 2 for SOS and
\item 3 for EOS .
\end{itemize}

The total number of tokens is 4, and the embedding vector
dimension is set to 8 which also becomes the Transformer model
dimension.

\paragraph{Loss function.} The MSE loss is not suitable for mapping decoder outputs to tokens. The mapping corresponds to classification and therefore Cross-Entropy loss is suitable. With the cross-entropy loss, a training scheduler becomes useful as well. A multi-step scheduler with an initial learning rate 0.01 and gamma value 0.1 was used. The current learning rate is multiplied by the gamma value after pre-defined numbers of epochs. Typically only one threshold, 1000 epochs, is needed for all the following experiments.   

\paragraph{Results.}
\texttt{TokenTransformer} learns to map sequences of zeros and ones to sequences of zeros and ones. Quite surprisingly, however, it cannot learn to produce an exact number of tokens. For example, the following outputs were produced:
\begin{verbatim}
Input sequence: [1, 1, 1, 1]
Output (predicted) sequence: [1, 1, 1, 1, 1, 1, 1, 1, 1, 1, 1, 1, 1, 1, 1]

Input sequence: [0, 0, 0, 0]
Output (predicted) sequence: [0, 0, 0, 0, 0, 0, 0, 0, 0, 0, 0, 0, 0, 0, 0]  
\end{verbatim}

To debug the reason for the failure to know when to stop, let's choose
another two types of sequences for which one works and another does not.

\paragraph{Training sequences.}
The following two cases, three-to-one and one-to-three,
help to understand in which cases \texttt{TokenTransformer} fails:
\begin{displaymath}
  \begin{split}
    \mathbf{X}: 0,0,0 \rightarrow \mathbf{Y}: 1\\
    \mathbf{X}: 1,1,1 \rightarrow \mathbf{Y}: 0
  \end{split}
\end{displaymath}
and
\begin{displaymath}
  \begin{split}
    \mathbf{X}: 1 \rightarrow \mathbf{Y}: 0, 0, 0\\
    \mathbf{X}: 0 \rightarrow \mathbf{Y}: 1, 1, 1
  \end{split} \enspace .
\end{displaymath}

\paragraph{Results.}
Similar to the previous example, 100 training samples were generated.
\texttt{TokenTransformer} was trained for 1000 epochs after which it achieved the final loss 0.0013 for the three-to-one and 0.4825 for the one-to-three case.

\texttt{TokenTransformer} learn correctly the three-to-one data:
\begin{verbatim}
Example 0
Input sequence: [1, 1, 1]
Output (predicted) sequence: [0]

Example 1
Input sequence: [0, 0, 0]
Output (predicted) sequence: [1]
\end{verbatim}
but cannot learn learn the one-to-three data:
\begin{verbatim}
Example 2
Input sequence: [1]
Output (predicted) sequence: [0, 0, 0, 0, 0, 0, 0, 0, 0, 0, 0, 0, 0, 0, 0]

Example 3
Input sequence: [0]
Output (predicted) sequence: [1, 1, 1, 1, 1, 1, 1, 1, 1, 1, 1, 1, 1, 1, 1]
\end{verbatim}

The problem with the one-to-three case seems to be similar to the previous data where input sequence should be repeated in the output.

\paragraph{Findings.}
The reason why \texttt{TokenTransformer} cannot learn one-to-three case, on
more generally many-to-many cases, is that inference is different from
training.

During training, \texttt{TokenTransformer} observes N input and M output tokens as
\begin{displaymath}
  \begin{split}
    &\mathbf{X}: SOS,1,EOS \rightarrow \mathbf{Y}: SOS,?, 0, 0,EOS\\
    &\mathbf{X}: SOS,1,EOS \rightarrow \mathbf{Y}: SOS,0, ?, 0,EOS\\
    &\mathbf{X}: SOS,1,EOS \rightarrow \mathbf{Y}: SOS,0, 0, ?,EOS\\
  \end{split} \enspace
\end{displaymath}
but since in inference the output tokens are generated one by one, the task is
\begin{displaymath}
  \begin{split}
    &\mathbf{X}: SOS,1,EOS \rightarrow \mathbf{Y}: SOS,?\\
    &\mathbf{X}: SOS,1,EOS \rightarrow \mathbf{Y}: SOS,0, ?\\
    &\mathbf{X}: SOS,1,EOS \rightarrow \mathbf{Y}: SOS,0, 0, ?\\
  \end{split} \enspace
\end{displaymath}

In other words, the transformer is trained to ``see the future'', but during inference the future does not exist (until it is generated), and therefore the transformer is confused. The reason why the three-to-one, or more generally many-to-one, case works, is that there is no future with a single output token and therefore inference and training become similar.

Future must be hidden from the transformer.

\subsection{Sequence masking}
To make the training and inference steps identical, the future output samples must be \textit{masked} during training so that they do not contribute to the decoder representation before they are observed.
In the \texttt{torch.nn.Transformer} implementation, the target outputs must be masked also during inference.

For output (target) masking, the \texttt{torch.nn.Transformer} class provides a static function \texttt{.generate\_square\_subsequent\_mask(seq\_len)} which gives the following output for the sequence length 5:
\begin{verbatim}
tensor([[0., -inf, -inf, -inf, -inf],
        [0., 0., -inf, -inf, -inf],
        [0., 0., 0., -inf, -inf],
        [0., 0., 0., 0., -inf],
        [0., 0., 0., 0., 0.]])
\end{verbatim}
The first row corresponds to the first iteration when only the first output is available for training or inference. For example, for the first inference round with the target output \texttt{2,1,1,1,3} all other tokens except \texttt{2} are masked. Then the predicted output is added to the list and all but \texttt{2, pred1} are masked in the second round. This continues until the EOS token is reached.

The \texttt{torch.nn.Transformer} implementation in the training and inference modes generates the same number of predictions as in the target outputs. The predicted sequence corresponds to the target output shifted one step to the right (toward the future).

\texttt{MaskedTokenTransformer} adds the masking operations to the \texttt{TokenTransformer}. This is done by adding support to the externally provided masks in the \texttt{forward()} function:
\begin{lstlisting}
class MaskedTokenTransformer(nn.Module):
       ...
      .forward(
       ...
       tgt_mask = None):

       ...
       out = self.transformer(src, tgt, tgt_mask=tgt_mask)
       ...
\end{lstlisting}

\paragraph{Results.}
After 2000 epochs \texttt{MaskedTokenTransformer} obtains the final loss 0.0654 and learns one-to-many and many-to-one sequences correctly, for example
\begin{verbatim}
Example 2
Input: [1]
Continuation: [0, 0, 0]

Example 3
Input: [0]
Continuation: [1, 1, 1]
\end{verbatim}

\paragraph{Training data.}
To further study the limits of \texttt{MaskedTokenTransformer} let's try with the following training sequences:
\begin{itemize}
  \item 0,1,0,1 $\rightarrow$ 0,1,0,1
  \item 1,0,1,0 $\rightarrow$ 1,0,1,0
\end{itemize}

Surprisingly, the model cannot learn these sequences.

\paragraph{Findings.}
Future masking solves the problems with previous training data examples, but the latest examples with the same symbols in different order cannot be solved.

The problem is central to the Transformer attention which correlates each sample with all others. The correlation value is the same between same tokens despite their position. Therefore it is not possible to make the internal representation aware of in which order the samples were observed. For example, internally the sequences 0-1-0-1, 1-0-1-0, and even 1-1-0-0, 0-1-1-0, and 0-0-1-1, all produce the same representation.

\subsection{Positional encoding}
The last element to make the previous Transformer structure learn all previous sequences, and many more, is the \textit{positional encoding}. Positional encoding adds information about the token position to each embedding vector.

\begin{figure}[h]
  \begin{center}
  \includegraphics[width=0.5\linewidth]{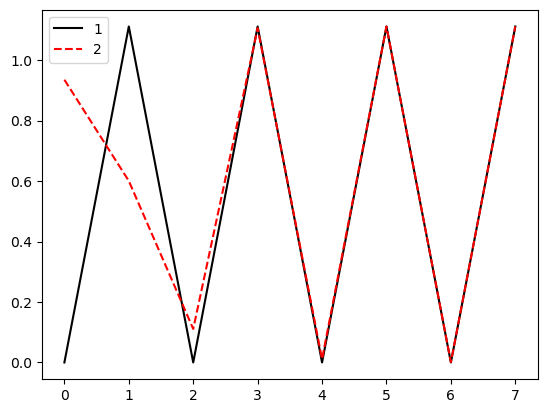}
  \caption{Positional encoding vectors for the position 1 (first) and 2 for 8-dimensional embedding vectors.}
  \label{fig:posencoding}
  \end{center}
\end{figure}

The adopted positional encoding is the one proposed in the original paper~\cite{transformer}. $\sin$ and $\cos$ functions are used to generate positional vectors added to token embedding vectors. Example position vectors are in Figure~\ref{fig:posencoding}.

Positional encoding is added to the transformer model after the token embedding:
\begin{lstlisting}
class Seq2SeqTransformer(nn.Module):
      ...
  .init(...)
      ...
      self.positional_encoder = PositionalEncoding(d_model=d_model, dropout=dropout_p)
      ...
      
   .forward(...):
      ...
      # Positional encoding - data must be seq len x batch num x feat dim
        # Inference often misses the batch num
        if src.dim() == 2: # seq len x feat dim
            src = torch.unsqueeze(src,1) 
        src = self.positional_encoder(src)
        if tgt.dim() == 2: # seq len x feat dim
            tgt = torch.unsqueeze(tgt,1) 
        tgt = self.positional_encoder(tgt)
       ...
\end{lstlisting}

\paragraph{Results.} The final Transformer model obtains the final loss of 0.0088 after 2000 epochs and outputs correct prediction:
\begin{verbatim}
Example 0
Input: [0, 1, 0, 1]
Continuation: [0, 1, 0, 1]

Example 1
Input: [1, 0, 1, 0]
Continuation: [1, 0, 1, 0]
\end{verbatim}

\paragraph{Findings.} The new model, \texttt{Seq2SeqTransformer}, has all the required elements implemented, and we demonstrated their importance with simple training sequences.

As the final step, all previous sequences are trained to the transformer to verify its ability to learn all of them at the same time.

\subsection{Token padding}
The final step is to teach all above sequences to a single transformer model:
\begin{itemize}
  \item 0,0,0,0 $\rightarrow$ 1,1,1,1
  \item 1,1,1,1 $\rightarrow$ 0,0,0,0
  \item   1,1,1 $\rightarrow$ 0
  \item   0,0,0 $\rightarrow$ 1
  \item       0 $\rightarrow$ 1,1,1
  \item       1 $\rightarrow$ 0,0,0
  \item 0,1,0,1 $\rightarrow$ 0,1,0,1
  \item 1,0,1,0 $\rightarrow$ 1,0,1,0
\end{itemize}
A problem the current model cannot handle is that sequences are of different lengths. There are two options, each sequence can be trained separately, which is inefficient, or dummy PAD tokens are added at the end of sequences that are shorter than the maximum length. If all sequences are approximately the same length, then the latter one is more efficient solution.

\paragraph{Padding masks.} Additional trouble with padding is that similar to masking of future tokens, the PAD tokens must be masked during training. There are three \texttt{torch.nn.Transformer.forward()} parameters in which the mask tensors must be provided:
\begin{itemize}
  \item Input masking (\texttt{src\_key\_padding\_mask})
  \item Output (target) masking (\texttt{tgt\_key\_padding\_mask})
  \item Decoder memory masking (\texttt{memory\_key\_padding\_mask})
\end{itemize}
In most of the cases decoder memory mask is the same as input mask, i.e. it prevents the decoder to see PAD tokens in its 'memory'.

\paragraph{Loss function.} A dummy embedding vector must be added for the padding token, and embedding informed that no gradient is computed for it:
\begin{lstlisting}
    self.embedding = nn.Embedding(num_tokens+1, d_model, padding_idx = self.padding_idx)
\end{lstlisting}

In addition, the loss function must be informed that errors in detecting the PAD tokens are ignored:
\begin{lstlisting}
  loss_fn = torch.nn.CrossEntropyLoss(ignore_index=PAD_IDX)
\end{lstlisting}

\paragraph{Results.} With 200 training samples (1/8 of each) and after 2000 epochs, the final loss is 0.1117 and \texttt{Seq2SeqTransformers} learns all Seq2Seq tasks correctly:
\begin{verbatim}
Example 0
Input sequence: [0, 0, 0, 0]
Output (predicted) sequence: [0, 0, 0, 0]

Example 1
Input sequence: [1, 1, 1, 1]
Output (predicted) sequence: [1, 1, 1, 1]

Example 2
Input sequence: [1, 1, 1]
Output (predicted) sequence: [0]

Example 3
Input sequence: [0, 0, 0]
Output (predicted) sequence: [1]

Example 4
Input sequence: [0]
Output (predicted) sequence: [1, 1, 1]

Example 5
Input sequence: [1]
Output (predicted) sequence: [0, 0, 0]

Example 6
Input sequence: [0, 1, 0, 1]
Output (predicted) sequence: [0, 1, 0, 1]

Example 7
Input sequence: [1, 0, 1, 0]
Output (predicted) sequence: [1, 0, 1, 0]
\end{verbatim}

\section{Conclusions}
The purpose of this work and accompanying Jupyter notebook for all examples is to familiarize one with the elements that are not part of the internal transformer structure but are essential to make it work. I hope you enjoy reading this article written in haste and the code written with great joy, and if you use any of them in your work I hope you cite this article in ArXiv.

\begin{ack}
This work would not be possible without my tenure in Tampere
University. This position allows me to allocate time to projects that interest me and sometimes I want to share my work and findings to thank the academic society and Finnish government. If you find this article or
related code useful, please cite this work (provide the ArXiv link, author name, and the article title in your Web page or article).

Universities all around the world are wonderful places for people like
me. I hope that the corporative thinking and style of management will
not destroy these institutes that have been developing for more than one
thousand years.

Keep on rockin' in the free world!
\end{ack}

\printbibliography

@inproceedings{Xiong-2020-icml,
      title={On Layer Normalization in the Transformer Architecture}, 
      author={Ruibin Xiong and Yunchang Yang and Di He and Kai Zheng and Shuxin Zheng and Chen Xing and Huishuai Zhang and Yanyan Lan and Liwei Wang and Tie-Yan Liu},
      year={2020},
      booktitle={International Conference on Machine Learning (ICML)},
      url={https://arxiv.org/abs/2002.04745},
}

@inproceedings{Cho-2014-emnlp,
    title = "Learning Phrase Representations using {RNN} Encoder{--}Decoder for Statistical Machine Translation",
    author = {Cho, Kyunghyun  and
      van Merri{\"e}nboer, Bart  and
      Gulcehre, Caglar  and
      Bahdanau, Dzmitry  and
      Bougares, Fethi  and
      Schwenk, Holger  and
      Bengio, Yoshua},
    booktitle = "Proceedings of the 2014 Conference on Empirical Methods in Natural Language Processing ({EMNLP})",
    year = "2014",
    url = "https://aclanthology.org/D14-1179/"
}

@inproceedings{Sutskever-2014-neurips,
      title={Sequence to Sequence Learning with Neural Networks}, 
      author={Ilya Sutskever and Oriol Vinyals and Quoc V. Le},
      booktitle = {Conference on Neural Information Processing Systems (NeurIPS)},
      year={2014},
      url={https://arxiv.org/abs/1409.3215}
}

@article{LSTM,
    author = {Hochreiter, Sepp and Schmidhuber, Jürgen},
    title = {Long Short-Term Memory},
    journal = {Neural Computation},
    volume = {9},
    number = {8},
    pages = {1735-1780},
    year = {1997},
    month = {11},
    url = {https://doi.org/10.1162/neco.1997.9.8.1735}
}

@book{MLBook,
    author = {C. Bishop},
    title = {Pattern Recognition and Machine Learning},
    publisher = {Springer},
    year = 2006
}

@inproceedings{transformer,
    author = {Ashish Vaswani and
                  Noam Shazeer and
                  Niki Parmar and
                  Jakob Uszkoreit and
                  Llion Jones and
                  Aidan N. Gomez and
                  Lukasz Kaiser and
                  Illia Polosukhin},
    title        = {Attention Is All You Need},
    booktitle = {Conference on Neural Information Processing Systems (NeurIPS)},
    year = 2017
}

\end{document}